\documentclass[10pt,twocolumn,letterpaper]{article}

\usepackage{iccv}
\usepackage{times}
\usepackage{epsfig}
\usepackage{graphicx}
\usepackage{amsmath}
\usepackage{amssymb}
\usepackage{multirow} 
\usepackage{booktabs} 

\newcommand*{\affaddr}[1]{#1} 
\newcommand*{\affmark}[1][*]{\textsuperscript{#1}}
\newcommand*{\email}[1]{\texttt{#1}}

\usepackage[pagebackref=true,breaklinks=true,letterpaper=true,colorlinks,bookmarks=false]{hyperref}

\newcommand{\minisection}[1]{\vspace{0.04in} \noindent {\bf #1}\ \ }

\iccvfinalcopy 

\def\iccvPaperID{4} 

\ificcvfinal\fi
\begin{document}

\title{Multi-Modal Fusion for End-to-End RGB-T Tracking}

\author{%
Lichao Zhang\affmark[1], Martin Danelljan\affmark[2], Abel Gonzalez-Garcia\affmark[1], Joost van de Weijer\affmark[1], Fahad Shahbaz Khan\affmark[3]\\
\affaddr{\affmark[1] Computer Vision Center, Universitat Autonoma de Barcelona, Spain}\\
\affaddr{\affmark[2] Computer Vision Laboratory, ETH Z\"urich, Switzerland}\\
\affaddr{\affmark[3] Inception Institute of Artificial Intelligence, UAE}\\
\email{\tt\small \{lichao,agonzalez,joost\}@cvc.uab.es, martin.danelljan@vision.ee.ethz.ch, fahad.khan@inceptioniai.org}\\
}

\maketitle
\thispagestyle{empty}

\begin{abstract}
We propose an end-to-end tracking framework for fusing the RGB and TIR modalities in RGB-T tracking. 
Our baseline tracker is DiMP (Discriminative Model Prediction), which employs a carefully designed target prediction network trained end-to-end using a discriminative loss. We analyze the effectiveness of modality fusion in each of the main components in DiMP, i.e. feature extractor, target estimation network, and classifier. 
We consider several fusion mechanisms acting at different levels of the framework, including pixel-level, feature-level and response-level. 
Our tracker is trained in an end-to-end manner, enabling the components to learn how to fuse the information from both modalities.
As data to train our model, we generate a large-scale RGB-T dataset by considering an annotated RGB tracking dataset (GOT-10k) and synthesizing paired TIR images using an image-to-image translation approach. We perform extensive experiments on VOT-RGBT2019 dataset and RGBT210 dataset, evaluating each type of modality fusing on each model component. The results show that the proposed fusion mechanisms improve the performance of the single modality counterparts. 
We obtain our best results when fusing at the feature-level on both the IoU-Net and the model predictor, obtaining an EAO score of 0.391 on VOT-RGBT2019 dataset. With this fusion mechanism we achieve the state-of-the-art performance on RGBT210 dataset. 
\end{abstract}

\vspace{-1mm}
\section{Introduction}
\vspace{-1mm}
As an important task in computer vision, visual object tracking, especially RGB tracking~\cite{bolme2010visual,henriques2015high,bertinetto2016fully,danelljan2014adaptive,danelljan2014accurate,danelljan2016beyond,danelljan2017eco,li2018high,zhu2018distractor,lukezic2017discriminative,danelljan2019atom}, has undergone profound changes in recent years. Researchers mainly focus on RGB tracking as large datasets are available~\cite{wu2015object,VOT_TPAMI,valmadre2018long}. However, RGB tracking obtains unsatisfactory performance in bad environmental conditions, e.g. low illumination, rain, and smog.
It was found that thermal infrared sensors provide a more stable signal for these scenarios. Therefore, RGB-T tracking has drawn more research attention recently~\cite{li2016learning,li2017weighted,li2018rgb,li2018cross}.

\begin{figure}[t]
\centering
    \includegraphics[width=\columnwidth]{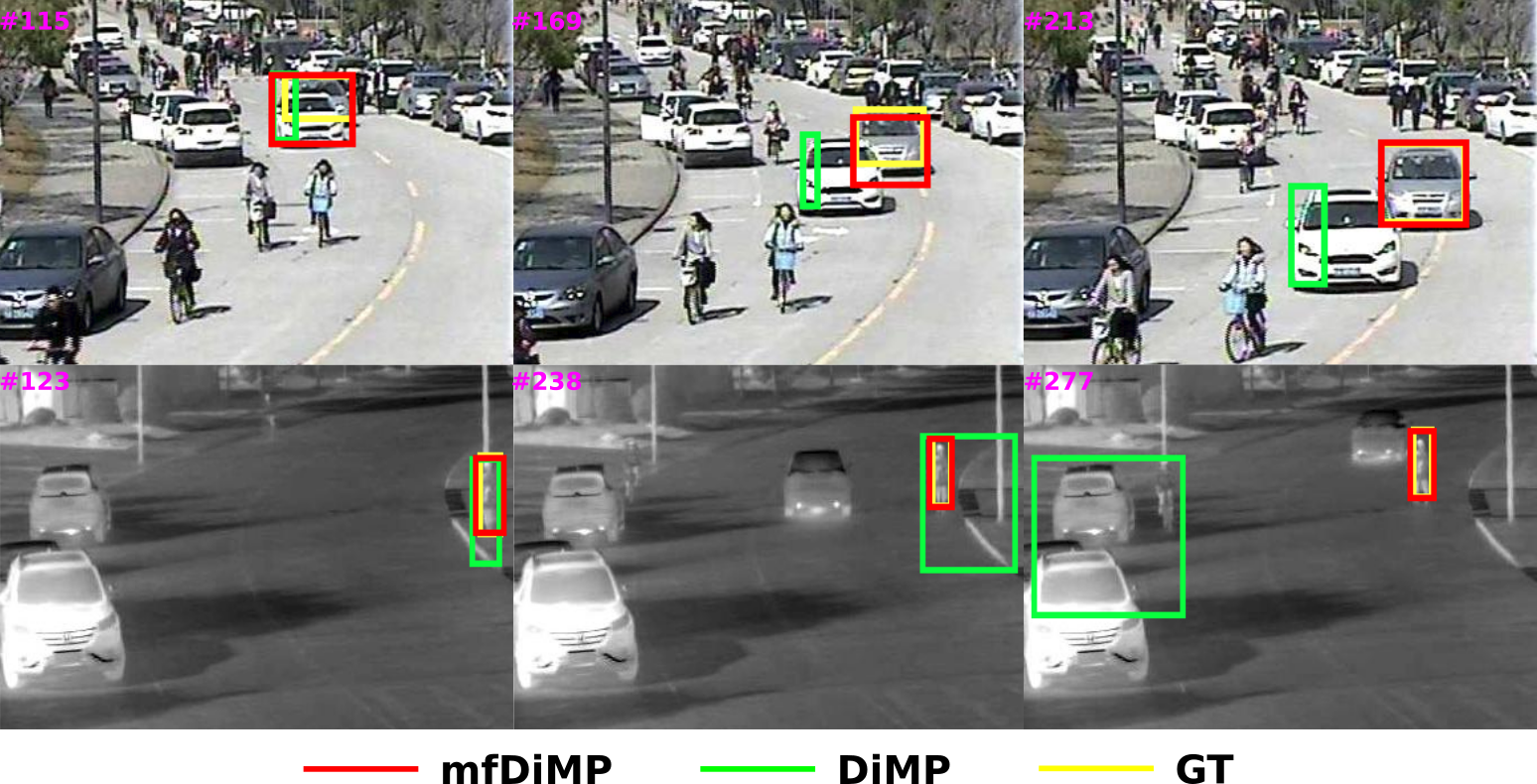}
    \caption{\small \textbf{Qualitative comparison between `mfDiMP' and `DiMP'.} Two exemplar videos from RGB modality and TIR modality on the top and bottom separately, where DiMP performs on each of them with single modality input. Our \emph{mfDiMP} can effectively track the object by fusing both modalities.}
    \vspace{-4mm}
    \label{fig:intro}
\end{figure}

As multi-modal data, i.e. from the RGB and TIR modalities, can provide complementary information for tacking, multi-modal tracking is a promising research direction. Images from the RGB modality have the advantage that they contain high-frequency texture information and provide rich representations for describing objects. Images from the TIR modality have the advantage that they are not influenced by illumination variations and shadows. Moreover, objects with elevated temperature can be distinguished from the background as the background is normally colder. Therefore, fusing the information from multi-modal data could benefit the tracker because it can exploit the complementary information of the modalities to improve tracking performance. 

There exists relatively little research on multi-modal tracking~\cite{wu2011multiple,liu2012fusion,li2016learning,li2017grayscale,li2018cross}. Most of these works are still using the sparse representation, normally with the hand-crafted features, for multi-modal tracking~\cite{liu2012fusion,li2016learning,li2017grayscale,li2018cross}. Later on in~\cite{li2018cross}, for comparison they design some baseline RGB-T trackers by extending the single modal tracker to a multi-modal tracker. 
This extension is done by directly concatenating the features from the RGB and the TIR modalities into a single vector, which is then fed into the tracker.
They also use some deep features for concatenation, but they are still off-the-shelf features pre-trained for other tasks. Therefore, there is still no previous work which investigates end-to-end training. We mention two main reasons for this. First, it is not obvious in what part of the tracking pipeline the fusion should be done. Ideally, we should fuse the information of the different modalities in such a way that it allows for optimal end-to-end training. Second, data scarcity of multi-modal tracking is a major obstacle to end-to-end training. Currently there are no large-scale aligned multi-modal datasets.
These two issues, i.e. no specific fusion scheme and lack of data, limit the progress of end-to-end multi-modal training.

To tackle this problem, in this paper we investigate how to effectively fuse multi-modal data in an end-to-end manner, which enables the optimal use of information from both modalities (see Figure~\ref{fig:intro}). We propose three end-to-end fusion architectures, consisting of pixel-level fusion, feature-level fusion, and response-level fusion. 
We use as baseline tracker the RGB tracker DiMP~\cite{bhat2019learning}.
To ensure that the proposed fusion tracker can be trained in an end-to-end manner, we also generate a large-scale paired synthetic RGB-T dataset with the method proposed in~\cite{zhang2019synthetic}. We perform extensive experiments on two commonly used benchmarks for RGB-T tracking: VOT-RGBT2019~\cite{vot2019} and RGBT210~\cite{li2017weighted}. Our multi-modal fusion tracker sets a new state-of-the-art on both datasets, achieving an EAO score of 0.391 on VOT-RGBT2019 and 55.5\% success rate on RGBT210. 

This paper is organized as follows. In section~\ref{RW}, we discuss the successful single modality tracking methods of recent years and the situation of current multi-modal tracking.
In section~\ref{DiMP}, we introduce the baseline tracker and analyze its components. In section~\ref{method}, we describe the proposed methods and formulations for the fusion of multi-modal tracking and provide the synthetic data for end-to-end training. In section~\ref{exp}, we present our extensive experiments on the VOT-RGBT2019 dataset and RGBT210 dataset. Finally, in section~\ref{conclusion}, we conclude our work and propose future research directions.
\vspace{-1mm}
\section{Related work}
\vspace{-1mm}
\label{RW}
\vspace{-1mm}
\subsection{Single modality tracking}
\vspace{-1mm}
Most current tracking algorithms focus on RGB images~\cite{bolme2010visual,henriques2015high,bertinetto2016fully,danelljan2014adaptive,danelljan2014accurate,danelljan2016beyond,danelljan2017eco,li2018high,zhu2018distractor,lukezic2017discriminative,danelljan2019atom}, although several approaches track in the TIR modality instead~\cite{zhang2019synthetic,yu2018online,yu2017dense,li2018cross}.
Despite the development of deep learning in many computer vision tasks, object tracking continued to use hand-crafted features during the first stage of deep learning~\cite{ma2015hierarchical,nam2016learning,danelljan2016beyond,danelljan2017eco,song2018vital}. Later on, some trackers~\cite{danelljan2016beyond,danelljan2017eco,ma2015hierarchical} pioneered in the involvement of deep features in tracking by using the pre-trained models for an image classification task~\cite{russakovsky2015imagenet}.
The main reasons for only using pre-trained models were the lack of large-scale training datasets and the difficulty of designing a suitable end-to-end training framework for tracking. Bertinetto \etal~\cite{bertinetto2016fully} proposed to train a network end-to-end by using a video object detection dataset~\cite{russakovsky2015imagenet}. Recently, several large-scale tracking datasets~\cite{fan2018lasot,muller2018trackingnet,huang2018got,valmadre2018long}, e.g.\ GOT-10k~\cite{huang2018got}, have been released with millions of images and various categories for training. Therefore, some current tracking approaches~\cite{danelljan2019atom,li2018siamrpn++,bhat2019learning} perform end-to-end training by leveraging these large-scale datasets. 

\minisection{RGB trackers.} 
Bertinetto \etal~\cite{bertinetto2016fully} proposed to use a fully-convolutional architecture to learn a similarity metric offline, i.e. a Siamese network. 
After training, the Siamese network is deployed for online tracking with high efficiency. To learn attention on the cross correlation, Wang \etal~\cite{wang2018learning} include additional attention components in the Siamese network and learn the spatial and channel weights for the \emph{exemplar} model. 
Li \etal~\cite{li2018high} utilize a proposal network to estimate the score maps and bounding boxes using two branches, which provides more accurate object scales than the traditional multi-resolution scale estimation. Later it is extended to use deeper and wider networks achieving significant improvement~\cite{li2018siamrpn++}.

An alternative approach to Siamese networks is correlation filter (CF) based tracking~\cite{bolme2010visual,henriques2015high,galoogahi2013multi,danelljan2014adaptive,zhang2014fast,zhang2019learning,zhang2016robust,danelljan2014accurate,danelljan2015learning,danelljan2016adaptive,cf_ca_tracking,kiani2017learning,lukezic2017discriminative,ma2015long}, which has occupied top positions for many years given its discriminative abilities and efficient tracking speed. The core part of CF trackers is the calculation of a filter that is later applied to detect the object in the search region of next frame. 
The calculation is performed in the Fourier domain, which makes it highly efficient. To overcome the issue of boundary effect in correlation filter in tracking, Danelljan \etal~\cite{danelljan2015learning} proposed to regularize the filter with a Gaussian window and Kiani \etal~\cite{kiani2017learning} proposed to use a mask formulated in the correlation filter. Some CF trackers~\cite{ma2015hierarchical,danelljan2015convolutional,danelljan2016beyond,danelljan2017eco} also benefited from pre-trained convolutional features.
CFNet~\cite{valmadre2017end} added end-to-end training by formulating CF as one layer of the network, although this only gives a marginal gain with respect to the baseline model, SiamFC~\cite{bertinetto2016fully}. Park \etal~\cite{park2018meta} proposed to learn an initial model for the correlation filter offline, accelerating the convergence speed for the filter optimization.

\minisection{TIR trackers.}
Contrarily to RGB tracking, most of the top performing TIR trackers still use hand-crafted features in their models.
For example, SRDCFir~\cite{felsberg2015thermal} extends the SRDCF~\cite{danelljan2015learning} tracker for TIR data by combining motion features with hand-crafted visual features, e.g. HOG~\cite{dalal2005histograms}, color names~\cite{van2009learning}, intensity, etc. 
EBT~\cite{zhu2016beyond} uses edge features to devise an objectiveness measure that generates high quality object proposals.
Yu \etal~\cite{yu2018online,yu2017dense} propose structural learning on dense samples around the object, using edge and HOG features~\cite{dalal2005histograms}, transferred to the Fourier domain for efficiency. 
Zhang \etal ~\cite{zhang2019synthetic} propose using an end-to-end trainable deep network.
They generate a large-scale TIR tracking dataset for training from existing RGB tracking dataset.
They use a current image translation approach~\cite{isola2017image} to synthesize a large amount of TIR images from RGB and they transfer the corresponding object annotations. 
By training the network with this data, they achieve state-of-the-art results in TIR tracking.
Following this idea, we obtain a large-scale RGB-T dataset that enables the use of deep learning for RGB-T tracking.

\vspace{-1mm}
\subsection{Modality fusion tracking}
\vspace{-1mm}

Fusing the RGB and TIR modalities is a promising direction. Some RGB-T trackers once have been proposed. Conaire \etal~\cite{conaire2008thermo} proposed to efficiently combine visible and thermal features by fusing the outputs of multiple spatiogram trackers, which is a derivation from mean-shift type algorithm~\cite{birchfield2005spatiograms}. Wu \etal~\cite{wu2011multiple} used a sparse representation for the target template by concatenating RGB and TIR image patches. Similarly, Liu \etal~\cite{liu2012fusion} also use a sparse representation by minimizing the coefficients from each modality. However, these methods provide sub-optimal fusions as both modalities contribute equally, while in practice one modality may have more valuable information than the other. Li \etal~\cite{li2016learning,li2017grayscale} addressed this with an adaptive fusion scheme to integrate visible and thermal information in the sparse representation by introducing weights to balance the contribution of each modality. In order to limit the effect of background clutter during tracking, Li \etal ~\cite{li2018cross} introduced a ranking between the two modalities, which is taken into account in the used patch-based features. They effectively avoided background effects by using the learned features with a structured SVM. 

As far as we know, all of the current RGB-T approaches use hand-crafted features, which significantly limits their tracking performance.
Although several RGB-T tracking datasets~\cite{li2016learning,li2017weighted,li2018rgb} have been recently released, they are only for testing purposes and are not large enough for training a deep learning based RGB-T tracker.
We propose adapting a deep RGB tracker for RGB-T by exploring different types of modality fusion, and performing end-to-end training with partly synthesized RGB-T data.

\begin{figure*}
    \centering
    \includegraphics[width=.95\textwidth]{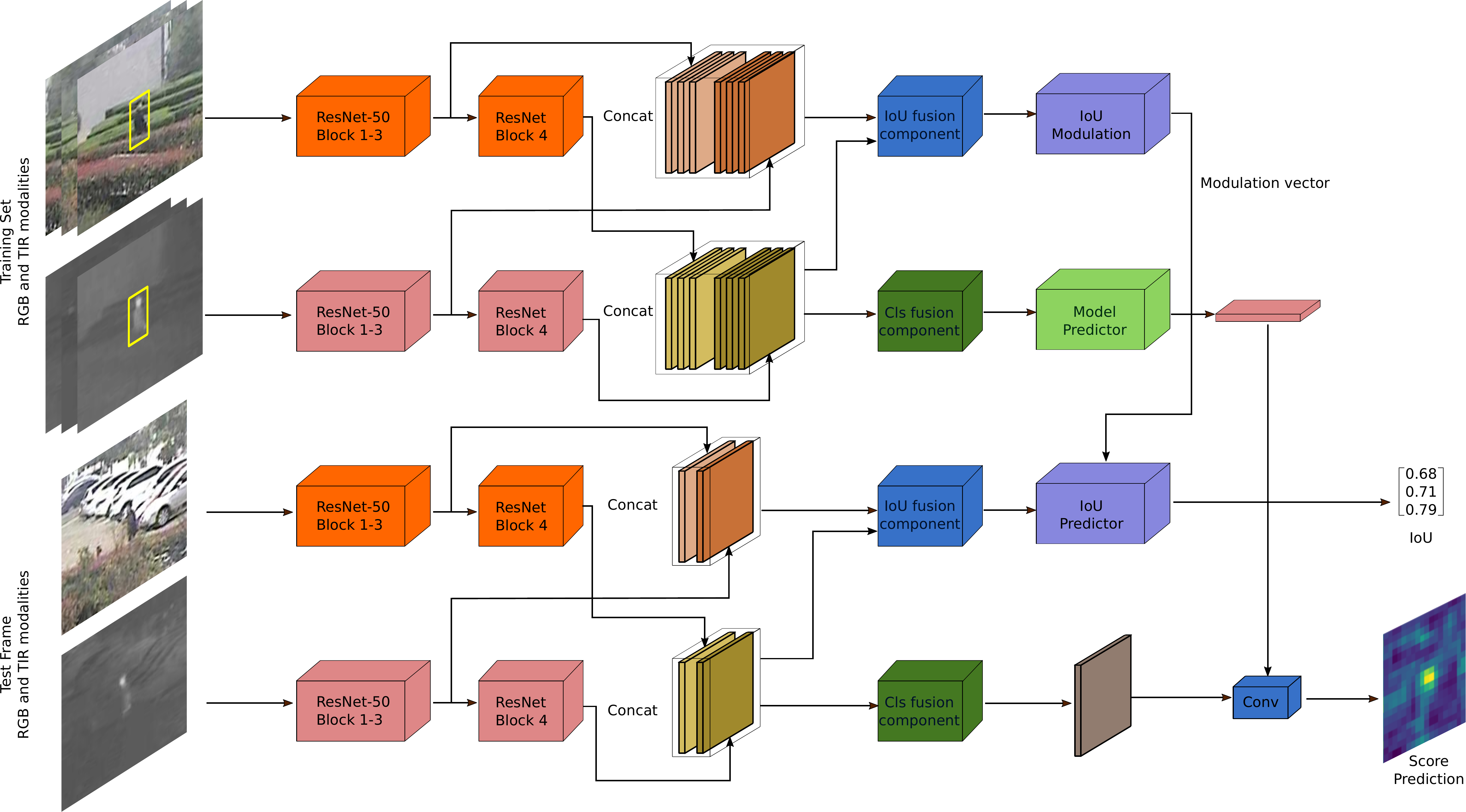}
    \caption{\small \textbf{Overview of our multi-modal fusion framework on feature-level.} We input images from RGB and TIR modalities to their feature extractor separately. Then we fuse the deep features from different blocks of the backbone. Fused features from block3 and block4 are input to IoU modulation and IoU predictor. Fused features from block4 are input to the model predictor for the final response map.} 
    \vspace{-4mm}
    \label{fig:framework}
\end{figure*}

\vspace{-1mm}
\section{Baseline RGB tracker}
\vspace{-1mm}
\label{DiMP}
In this section, we describe the architecture of the tracker we have selected for our multi-modal tracking experiments. We use the Discriminative Model Prediction (DiMP) tracker~\cite{bhat2019learning}, which was originally proposed for single modality tracking.

\minisection{Discriminative Model Prediction.}
\label{sec: DiMP}
DiMP~\cite{bhat2019learning} proposed an end-to-end trainable tracking architecture, capable of learning a powerful discriminative filter by embedding the online learning of the target model into itself. DiMP consists of the following components: feature extractor, model predictor, and target estimation network (IoU-Net~\cite{jiang2018acquisition}). With these carefully designed components and an effective optimization method, they achieve excellent performance on RGB tracking by setting a new state-of-the-art on several RGB tracking datasets~\cite{kristan2018sixth,wu2015object,huang2018got,muller2018trackingnet,fan2018lasot}.

\minisection{Feature extractor.}
The backbone feature extractor $F$ normally aims to extract the deep feature representations for the follow-up implementation models. Here, specifically in DiMP~\cite{bhat2019learning}, the deep representations are extracted for the model predictor and target estimation network.

DiMP~\cite{bhat2019learning} employs the ResNet-18 and ResNet-50 architectures, which is trained on ImageNet, as the backbone feature extractors for DiMP-18 and DiMP-50 separately. They implement fine-tuning the backbone for the end-to-end training. After an analysis on the impact of different feature blocks in DiMP~\cite{bhat2019learning}, they use the features from block3 and block4 for IoU-Net, and only from block4 for the classifier. The feature extractor $F$ is shared and only performed on a single image patch per frame.

For training the feature extractor $F$, they input data for $F$ with a pair of sets $(M_{train},M_{test})$. Each set $M=\{(I_j,b_j)\}_{j=1}^{N_{frames}}$ contains images $I_j$ along with their object bounding box $b_j$. The target model is predicted by using $M_{train}$ and then evaluated on the test frames $M_{test}$. $M_{train}$ and $M_{test}$ are constructed by sampling $N_{frames}$ frames for both from first and second halves of the segment respectively.
They pass the images through the feature extractor $F$, and obtain the train set $S_{train}=\{(x_j,c_j)\}$, where $x_j=F(I_j)$, and $c_j$ is the center coordinate of the box $b_j$. 

\minisection{Model predictor and response map.}
The Model predictor is to obtain the final optimized filter $f$, which consists of model initializer, which is a convolutional layer followed by a precise ROI pooling\cite{jiang2018acquisition}, and model optimizer, which is to solve the final model $f$ by the steepest descent (SD).
The model filter $f$ is solved by  using multiple samples in $S_{train}$, which happens in the model initializer.
The input of the model predictor is a set of $S_{train}$, and obtain the model $f$ by training on the model predictor: $f=D(S_{train})$. Then the filter $f$ is evaluated on the test samples $S_{test}$ and finally classification loss for offline training is computed as:
\begin{equation}
    \vspace{-1mm}
    L_{cls}  = \frac{1}{N_{iter}}\sum_{i=0}^{N_{iter}}\sum_{(x,c)\in\ S_{test}} \left\|l(x*f^{(i)},z_c) \right\|^2.
    \vspace{-1mm}
    \label{eq:loss} 
\end{equation}
Where, $z_c$ is a Gaussian function centered as the target $c$. $f^{(0)}$ is the output of model initializer. The response map can be calculated as: $s = x*f, x\in S_{test}$.

\minisection{Bounding box estimation.}
DiMP uses an IoU-Net based architecture from the ATOM tracker~\cite{danelljan2019atom}. The function of the IoU-Net model is to predict the IoU between the deep feature $x$ of an image and a bounding box candidate $B$. Bounding box estimation is then performed by maximizing the IoU prediction.
The network has two branches, one is the \emph{IoU modulation} for calculating the modulation vector from reference image, and the other branch is \emph{IoU predictor} for predicting the IoU values from test image. Then the reference branch is added with a convolutional layer, while the test branch is added with two convolutional layers as it dominates the IoU prediction. Both of them then are followed by PrPool (Precise ROI Pooling)\cite{jiang2018acquisition} and a fully connected layer. Here the interaction between the two branches is that a precomputed vector in the reference branch is used to modulate the feature representation of the test image via channel-wise correlation. The IoU is predicted in terms of the bounding box $B$ as follows:
\begin{equation}
    \vspace{-1mm}
    IoU(B)=g(c(x_0,B_0) \cdot z(x,B))
    \vspace{-1mm}
    \label{eq:iou} 
\end{equation}
Where, $x_0,B_0$ are from the reference image, and $x,B$ are from the test image. $z$ is the feature representation after PrPool layer in test branch. $g$ is the IoU predictor with three fully connected layers. $c$ is a modulation vector. 
\vspace{-1mm}
\section{End-to-end multi-modal tracking}
\vspace{-1mm}
\label{method}

There are two main issues when extending state-of-the-art RGB trackers to multi-modal data such as RGB-T.  First, a fusion component is not considered as a native designed component for the RGB tracker architecture, since the tracker only considers a single modality as input. 
Therefore, when extending to multi-modal data, these trackers must be equipped with a fusion strategy. 
Second, the lack of large-scale paired RGB-T training datasets 
complicates the end-to-end training of feature representations, which have been shown to significantly improve results for RGB tracking. 
To tackle the former, we investigate how to effectively fuse multi-modal data for tracking, aiming to make the best use of all available data modalities, in this case, RGB and TIR.
To tackle the latter, we ensure that the proposed multi-modal tracker can be trained in an end-to-end manner by generating a large-scale paired synthetic RGB-T dataset, similarly to the method proposed in~\cite{zhang2019synthetic}.

In this section, we first comprehensively explain our three end-to-end multi-modal fusion architectures, namely pixel-level fusion, feature-level fusion, and response-level fusion. 
We also explain how we apply~\cite{zhang2019synthetic} to generate a large multi-modal dataset.

\vspace{-1mm}
\subsection{Multi-modal fusion for tracking}
\vspace{-1mm}
\label{fusion}
In this subsection, we investigate three different mechanisms for multi-modal fusion with the aim to find the optimal fusion architecture.
We start the fusion work on the input of the network (pixel-level). Then we explore the fusion on the intermediate of the network. 
In~\cite{li2018cross}, they extended some RGB trackers by concatenating the RGB and TIR features into a single vector, and then used them as off-the-shelf features for the classifiers of various trackers. 
In contrast, we end-to-end train fused features, which are input to both the model predictor and the target estimation network (feature-level). 
Moreover, we explore fusion on the final response maps of the network (response-level).

\vspace{-1mm}
\minisection{Pixel-level fusion.} 
The first modality fusion we consider is at the input of the network. We propose to fuse the RGB and TIR images by directly concatenating the images along the channel direction and then inputting the fused RGB-T image to the feature extractor. To complete this fusion, we extend the filter size of the first layer in feature extractor from $7\times 7\times 3 \times 64$ to $7\times 7\times 4 \times 64$. The images that are input to the feature extractor should be concatenated as  $I^{F}=[I^V | I^T]$, where $I^V$ is the RGB image, $I^T$ is the TIR image and the fused image is $I^F$. 

\vspace{-1mm}
\minisection{Feature-level fusion.} 
To delay the fusion to a more semantically aware network stage, we evaluate the fusion effectiveness in the intermediate part of the network architecture.
Concretely, we implement the fusion after the feature extractor, i.e. fusing the deep feature representations from the RGB and TIR modalities.
We pass the RGB and TIR images through the feature extractors separately and extract features from both modalities independently. 
Then, we concatenate the features from each modality and feed them into the IoU predictor and model predictor. 
This provides a more expressive representation for the IoU predictor and more discriminative features for the model predictor. 
The framework of our proposed method for multi-modal fusion on feature-level is shown in Figure~\ref{fig:framework}, where we show how we concatenate the feature representations output by the feature extractors. 
The feature concatenation can be expressed as intuitive syntax: $x^{F}=[x^V | x^T]$. Here, $x^V$ is the features from the RGB modality, $x^T$ is the features from the TIR modality, and $x^{F}$ is the fused features. 

\vspace{-1mm}
\minisection{Response-level fusion.} 
To evaluate the effectiveness of an ensemble of independently trained trackers on each modality, we perform the multi-modal fusion on the final part of the training architecture in DiMP~\cite{bhat2019learning}, i.e. response-level fusion. For the response-level fusion, we use a pair of feature extractors and model predictors to process each image from RGB and TIR modalities separately. Finally, we sum their response maps to get the fused response map. 
We input a single modality to the IoU-Net component and thus there are two cases for training the whole network, one using the RGB modality to fine-tune IoU-Net and one using the TIR modality instead. 
Assuming that we have two single modality response maps, $s^V$ from the RGB modality and $s^T$ from the TIR modality, we calculate the fused response map by summing them: $s^{F} = s^V + s^T$.

\vspace{-1mm}
\subsection{RGB-T data generation} 
\vspace{-1mm}
The lack of large-scale paired RGB-T training datasets hampers end-to-end tracking in RGB-T datasets. We borrow the method from~\cite{zhang2019synthetic}, which proposes to use image-to-image translation methods to generate synthetic TIR data for tracking. 
In their paper, they show that using such data improves results for end-to-end training of TIR trackers. 
Here we explain how we generated the training data aiming for fine-tuning the pre-trained DiMP models. We take advantage of a normal RGB training dataset for RGB tracking, and then generate the TIR images by a well-trained image-to-image translation model~\cite{isola2017image}. With the above steps, we obtain an aligned synthetic RGB-T training dataset for RGB-T tracking.
As a result, our proposed fusion architectures (see section~\ref{fusion}) can also benefit from end-to-end training. 
Ideally, this will allow us to obtain a performance gain proportional to that observed for RGB tracking. 

After applying the described process, we obtain two datasets for RGB-T tracking, $(M_{train},M_{test})$, both of the form $M=\{(I_j^V,I_j^T,b_j)\}_{j=1}^{N_{frames}}$. Here, $I_j^T$ represents the $j$-th synthetic TIR image generated from the aligned RGB image $I_j^V$, and $b_j$ is their identical bounding box.

\begin{table*}[t]
    \begin{center}
    \resizebox{.9\textwidth}{!}{
        \begin{tabular}{l|cccc|ccc} 
            \hline
 Fusion level & Feature extractor & IoU-Net & Model predictor & Response map & EAO ($\uparrow$)& A ($\uparrow$)& R ($\downarrow$)\\  \hline
\multirow{6}{*}{Single modality}& RGB & RGB &RGB &RGB  & 0.327&0.586&0.345 \\
& TIR & TIR  &TIR &TIR   & 0.331&0.584&0.332 \\
& TIR & TIR (ft) &TIR (ft) & TIR    & 0.336&0.587&0.331 \\
& TIR (ft) & TIR  &TIR (ft) &TIR    & 0.339&0.589&0.329 \\
& TIR (ft) & TIR (ft) &TIR (ft) &TIR    & 0.341&0.590&0.328 \\ 
& RGB (ft) & RGB (ft) &RGB (ft) &RGB    & 0.335&0.586&0.331 \\ \hline
Pixel-level   & RGB+TIR (ft)  & RGBT (ft)&RGBT (ft) & RGBT  &0.345 &0.552 & 0.281 \\ \hline
\multirow{2}{*}{Response-level}& RGB/TIR (ft)& RGB (ft) &RGB/TIR (ft) & RGB+TIR  & 0.342 &0.546 & 0.309 \\
& RGB/TIR (ft)& TIR (ft) &RGB/TIR (ft) & RGB+TIR  &0.349 &0.554 & 0.291 \\ \hline
\multirow{6}{*}{Feature-level}& RGB/TIR (ft)&RGB (ft) &RGB+TIR (ft) & RGBT &0.346 &0.545 &0.266   \\
& RGB/TIR (ft)&TIR (ft) &RGB+TIR (ft) & RGBT  &0.359 &0.564 & 0.243\\
& RGB/TIR (ft)&RGB+TIR (ft) &RGB (ft) &RGB  &0.354 &0.563 & 0.276 \\
& RGB/TIR (ft)&RGB+TIR (ft) &TIR (ft) &TIR  &0.366 &0.601 & 0.261 \\
& RGB/TIR (ft)&RGB+TIR (ft) &RGB+TIR (ft) &RGBT  &0.389 &0.605 & \textbf{0.224} \\
& RGB/TIR (ft / ft$\times10$)&RGB+TIR (ft) &RGB+TIR (ft) &RGBT &\textbf{0.391} &\textbf{0.615}&0.228 \\ \hline         
        \end{tabular}}
        \vspace{1mm}
        \caption{\small \textbf{Fusion mechanisms analysis on VOT-RGBT2019~\cite{vot2019}.} We evaluate several fusion mechanisms at different levels of DiMP~\cite{bhat2019learning}. The results are reported in terms of EAO, normalized weighted mean of accuracy (A), and normalized weighted mean of robustness score (R). We explicitly show the input modality for each component of the tracker. Here, `RGB' and `TIR' are the single modality,`RGB/TIR' means each modality input separately, `RGB+TIR' means that both modalities are input simultaneously, and `RGBT' indicates fused features from both modalitites used in the remaining of network. Finally, (`ft') means fine-tuning and (`ft$\times10$') means fine-tuning with a higher learning rate. The best results are highlighted in bold font.}
        \vspace{-6mm}
        \label{table:components analisis}
    \end{center}
    
\end{table*}

\vspace{-1mm}
\section{Experiments}
\vspace{-1mm}
\label{exp}
In this section, we provide a comprehensive evaluation of the proposed tracker mfDiMP on two benchmarks, VOT-RGBT2019~\cite{vot2019} and RGBT210~\cite{li2017weighted}, and describe all implementation and evaluation details.

\vspace{-1mm}
\subsection{Generating the training RGB-T dataset}
\vspace{-1mm}
\label{sec:training_dataset}

We use the recent Generic Object Tracking Benchmark (GOT-10k)~\cite{huang2018got} to train our fused modality networks. 
GOT-10k has over 10,000 video segments, covering 563 classes of real-world moving objects and more than 80 motion patterns, amounting to a total of over 1.5 million manually labeled bounding boxes.
It also provides additional supervision in terms of attribute labels such as ratio of object visible or motion type.
We employ GOT-10k's training set, which contains 9,335 videos (1,403,359 frames), with 480 object classes and 69 motion classes.
We refrain from using the set of 1000 prohibited videos listed in the VOT challenge website~\cite{vot2019}, so we train our model with the remaining 8,335 videos (1,251,981 frames).

With this reduced version of GOT-10k RGB dataset, we generate a large-scale RGB-T dataset by synthesizing paired TIR images using an image-to-image translation approach, as in~\cite{zhang2019synthetic}. 
Specifically, we use pix2pix~\cite{isola2017image} for image-to-image translation given its excellent performance~\cite{zhang2019synthetic}. 
To train the pix2pix model, we use a total of 87K pairs of aligned images in the RGB and TIR modalities, depicting several different scenarios. 
These images are carefully collected and arranged from many current existing RGB-TIR datasets~\cite{zhang2019synthetic}.
We train the pix2pix model using the default settings described in~\cite{zhang2019synthetic}. 
After training, we use pix2pix to transfer the selected RGB videos in GOT-10k~\cite{huang2018got} to synthetic TIR videos, along with the labels.

\vspace{-1mm}
\subsection{Evaluation datasets and protocols}
\vspace{-1mm}

\minisection{VOT-RGBT2019 dataset~\cite{vot2019}} contains 60 public testing sequences, with a total of 20,083 frames. It is used as the most recent edition of the VOT challenge.
We follow the VOT protocol, which establishes that when the evaluated tracker fails, i.e.\ when the overlap with the ground-truth is below a given threshold, it is re-initialized in the correct location five frames after the failure. 
The main evaluation measure used to rank the trackers is Expected Average Overlap (EAO), which is a combination of accuracy (A) and robustness (R). 
We compute all results using the provided toolkit~\cite{vot2019}.

\minisection{RGBT210 dataset~\cite{li2017weighted}} contains 210 highly-aligned public RGB and TIR video pairs for testing, with 210K frames in total and a maximum of 8K frames per sequence pair. 
There are a total of 12 representative attributes, such as camera moving, large scale variations and environmental challenges, which are annotated for each video. {These facilitate attribute-sensitive evaluation analyses.} We compare our results with other trackers using the provided toolkit~\cite{lirgbtdataset}.
We use precision plot and success plot to evaluate the trackers.

\begin{figure*}[th!]
    \centering
    \includegraphics[width=.75\textwidth]{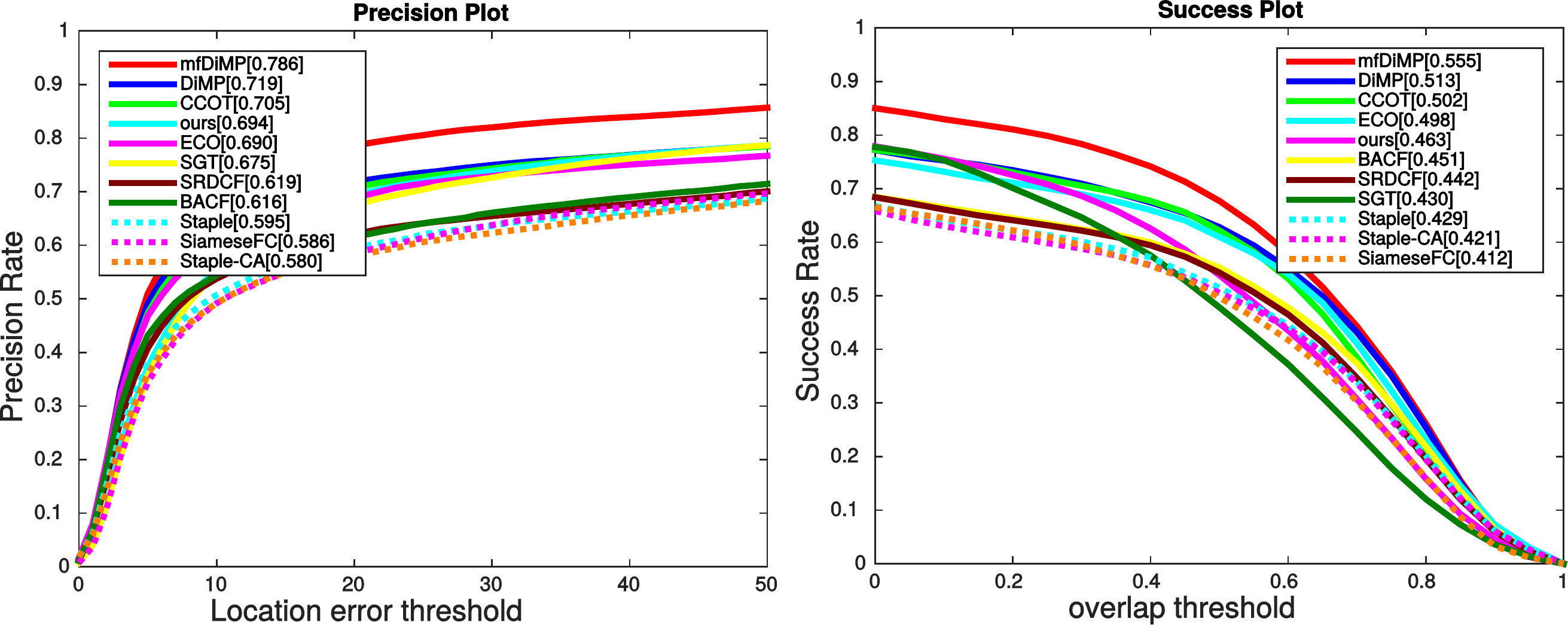}
    \caption{\small \textbf{Precision plot and success plot by comparing our mfDiMP with the top-10 trackers on RGBT210 dataset~\cite{li2017weighted}}. We can see our mfDiMP outperforms DiMP with an absolute gain of 6.7\% and 4.2\% in terms of precision rate and success rate respectively.}
    \vspace{-5mm}
    \label{fig:RGBT210_plot}
\end{figure*}
\vspace{-1mm}
\subsection{Implementation details}
\vspace{-1mm}
We use DiMP~\cite{bhat2019learning} as our base tracker with ResNet-50~\cite{he2016deep} as backbone network.
The base architecture of DiMP is pre-trained on several large-scale RGB training datasets~\cite{fan2018lasot,muller2018trackingnet,huang2018got,lin2014microsoft}. 
To test the single modality versions, we simply input the images from either modality as in traditional RGB trackers~\cite{henriques2015high,danelljan2017eco,bertinetto2016fully}.
RGB images have 3 image channels while TIR images have 1 channel, and so the pixel-level fusion uses 4-channel images. For the feature-level fusion, we concatenate the convolutional features after the feature extractors. Finally, for the response-level fusion we add together the final confidence maps independently predicted by the RGB and TIR modalities.

We use separate, modality-specific feature extractors for the response-level fusion and feature-level fusion. As hyperparameters for fine-tuning our architecture, we use the default values used to train each component in DiMP~\cite{bhat2019learning}, which have been carefully set and described by the authors in section 3.2 of~\cite{bhat2019learning}. 
We keep the default learning rates for each component as in the DiMP model and then decrease them by collaboratively multiplying a small gain learning rate, i.e. 0.001 when fine-tuning. 
In one of our experiments, we set the learning rate for the TIR feature extractor $\times10$ higher than for the RGB feature extractor. 
We do this considering that the RGB feature extractor was pre-trained with a large-scale RGB dataset, leading to satisfactory RGB features. 
On the other hand, the TIR feature extractor needs to catch up with that of the RGB modality in terms of learning, and thus it requires a higher learning rate for fine-tuning.

As a result of the stochastic nature of DiMP, the tracker generates different results for every run.
Following the procedure employed in~\cite{bhat2019learning}, we compute the default 15 runs of our mfDiMP tracker for VOT-RGBT2019 dataset and 5 for RGBT210 dataset. %
Then we obtain the final result by averaging the results of all runs.

\vspace{-1mm}
\subsection{Analysis of fusion mechanisms}
\vspace{-1mm}
Table~\ref{table:components analisis} presents our analysis  to determine the best location to fuse the modalities in DiMP.
The table is an extensive evaluation of all considered fusions under different configurations.
We start with a comprehensive evaluation of the baseline tracker DiMP~\cite{bhat2019learning} for single modality. 
We present several configurations in the upper part of Table~\ref{table:components analisis} (`Single modality') using the RGB modality or TIR modality  alone. 
The first two rows are the original DiMP (pre-trained for RGB) using either RGB or TIR images during online tracking.
We observe how the TIR images obtain a higher result. 
For the next three rows, we fine-tune the feature extractor and/or IoU-Net with synthetic TIR images. 
In this case, fine-tuning the single modality network for TIR improves the pre-trained networks with an absolute gain of 1\%. 
Finally, we can see how fine-tuning only on RGB improves the performance of the pre-trained model, but to a lesser extent than using TIR.
In the lower part of Table~\ref{table:components analisis} we analyze the effectiveness of each fusion mechanism for DiMP~\cite{bhat2019learning}, which we discuss in detail in the remainder of this section.

\vspace{-1mm}
\minisection{Pixel-level fusion.}
From Table~\ref{table:components analisis}, we can observe that pixel-level fusion improves the performance of the baseline tracker from 0.331 to 0.345 with an absolute gain of 1.4\%. 
The images from the RGB and TIR modalities have complementary information. 
Therefore, using the fused images to train the network end-to-end 
can help the model to learn better deep feature representations, which in turn improves tracking performance.

\begin{table}[t]
    \begin{center}      
    \vspace{2mm}
    \resizebox{\columnwidth}{!}{
        \begin{tabular}{ccccccc}
        \hline
 & ECO & SiamFC & DaSiamRPN & ATOM & DiMP & \multirow{2}{*}{\textbf{mfDiMP}}\\ 
 & \cite{danelljan2017eco} & \cite{bertinetto2016fully} & \cite{zhu2018distractor} & \cite{danelljan2019atom} &  \cite{bhat2019learning} & \\ \midrule
EAO($\uparrow$) &  0.265 & 0.254 &0.324  & 0.318 & 0.327 & {\textbf{0.391}} \\ \hline
A ($\uparrow$) &  0.580  & 0.594 & 0.604 & 0.575& 0.586 & \textbf{0.615}\\ \hline
R ($\downarrow$)  & 0.480 &0.533 & 0.482 & 0.374 & 0.345 & \textbf{0.228}\\ \hline
FPS ($\uparrow$) &  11.2  & 38.1 & \textbf{62.4} & 12.1& 13.6 & {10.3}\\ \hline
        \end{tabular}}
        \vspace{0mm}
         \caption{\small{\textbf{State-of-the-art comparison on VOT-RGBT2019 dataset.} Our mfDiMP improves the baseline tracker DiMP with an absolute gain of 6.4\% in terms of EAO {without significantly deteriorating the computational efficiency}. The best results are highlighted in bold font.}}
         \vspace{-6mm}
        \label{table:VOTRGBT2019}
    \end{center}
\end{table}

\vspace{-1mm}
\minisection{Response-level fusion.}
In this case, the fusion takes place in the final response map output by the classifier. 
The signals from both modalities pass through the classifier separately and both compute the response map. Then we sum together the two response maps, obtaining the final fused response map.
Meanwhile, we input a single modality for the IoU-Net, either RGB or TIR.
Both fusion mechanisms enhance the tracking performance, and using the TIR modality for IoU-Net outperforms using RGB, achieving scores of 0.349 and 0.342, respectively. 
The results show that fusion on response-level obtains the same effectiveness as pixel-level fusion.

\begin{table*}[t!]
    \begin{center}      
    \resizebox{.8\textwidth}{!}{
        \begin{tabular}{l|c|c|c|c|c|c|c}
        \hline
 & {ECO}& {CSR} &  {CMRT} &  SGT &CNN+KCF+&CFnet+ & \multirow{2}{*}{\textbf{mfDiMP}}\\  
& \cite{danelljan2017eco}& \cite{lukezic2017discriminative} &\cite{li2018cross}   &\cite{li2017weighted} &RGBT\cite{henriques2015high}&RGBT\cite{valmadre2017end} &  \\\hline
No Occlusion        &87.7/64.3 &68.1/45.2  & 86.1/59.4   & 82.4/50.7&63.7/42.9 &69.7/52.2& \textbf{88.9/67.3}\\ 
Partial Occlusion   &72.2/52.5 &52.7/36.6  & 77.1/52.2   & 75.4/48.3&56.0/36.4 &57.2/38.4& \textbf{84.0/60.1}\\ 
Heavy Occlusion     &58.3/41.3 &37.1/24.3  & 54.6/34.8   & 53.1/34.1&36.6/25.9 &39.3/27.3& \textbf{68.4/45.8}\\
Low Illumination    &66.6/45.6 &47.3/31.1  & 71.4/46.4   & 71.6/44.7&52.8/34.5 &49.8/33.6& \textbf{77.1/53.7} \\
Low Resolution      &64.1/38.1 &46.0/23.1  & 64.7/37.4   & {65.8}/37.5&54.6/32.5 &45.2/27.7& \textbf{69.2/43.6}\\
Thermal Crossover   &\textbf{82.1/58.8} &43.2/29.3  & 65.8/43.0   & 64.9/40.7&49.6/33.2 &42.8/29.4& 76.5/55.2\\
Deformation         &61.2/45.0 &44.7/33.0  & 65.2/45.8   & 65.3/45.9&44.8/34.4 &48.9/35.2& \textbf{77.7/56.6}\\
Fast Motion         &58.2/39.2 &42.6/25.0  & 58.8/34.9   & 58.0/33.1&37.1/24.1 &36.5/23.0& \textbf{76.7/52.6}\\
Scale Variation     &74.5/55.4 &53.3/37.5  & 72.5/49.2   & 67.4/41.7&50.3/32.6 &56.7/40.6& \textbf{82.2/59.5}\\
Motion Blur         &67.8/49.9 &34.7/23.8  & 58.4/40.5   & 58.6/39.6&30.4/22.0 &30.3/22.4& \textbf{72.5/51.2}\\
Camera Moving       &61.7/45.0 &38.9/27.4  & 60.0/41.9   & 59.0/40.7&36.2/27.0 &37.2/27.9& \textbf{75.3/53.8}\\
Background Clutter  &52.9/35.2 &38.4/23.7  & 58.3/35.6   & 58.6/35.5&42.3/28.4 &43.7/28.1&\textbf{71.5/45.7}\\\hline
ALL &69.0/49.8 &49.1/33.0  & 69.4/46.3   & 67.5/43.0&49.3/33.1 &51.8/36.0& \textbf{78.6/55.5}\\ \hline
        \end{tabular}}
        \vspace{1mm}
         \caption{\small{\textbf{Attribute-based Precision Rate and Success Rate (PR/SR \%) on RGBT210 dataset with several trackers.} These trackers include popular RGB trackers such as ECO and CSR, recent multi-modal fusion tracker like CMRT and SGT, and also extended RGB-T trackers from KCF and CFnet. Our tracker surpasses almost all the trackers over all the attributes.}}
         \vspace{-6mm}
        \label{table:RGBT210_attributes}
    \end{center}
\end{table*}

\vspace{-1mm}
\minisection{Feature-level fusion.}
We consider inputting fused feature representations into two different DiMP components, i.e. model predictor and IoU-Net. 
In the former case, only the model predictor receives fused features, whereas the remaining component (IoU-Net) still uses features from a single modality. 
In both cases, we improve over single modality tracking, and the version with TIR for IoU-Net obtains a better result with 0.359, an absolute gain of 2.8\% with respect to the best single modality model (fine-tuned on TIR, row 5 of Table~\ref{table:components analisis}). 
We obtain an even higher score, 0.366, when fusing features for IoU-Net while using TIR features for the model predictor. 
This demonstrates that using fused features for IoU-Net outperforms using fused features for the model predictor. 
We attribute this to the fact that IoU-net is a more complex network than the model predictor, where the fused feature representations can be more effective to prompt IoU-Net to its ultimate capacity.
Finally, we use fused features to feed both model predictor and IoU-Net, which significantly improves the result to 0.389, a substantial absolute gain of 5.8\%.
Considering that the feature extractor is pre-trained on RGB images, it is natural to assume that the feature extractor for TIR images needs a stronger training signal. 
For this reason, we propose a variant in which the feature extractor for the TIR modality has a higher learning rate ($\times10$). 
This variant achieves 0.391, which is the best result and significantly outperforms the best single modality tracker with a big jump. 

Form Table~\ref{table:components analisis}, we can see that fusion on feature-level with end-to-end training provides significant improvements in tracking performance. 
Specifically, fusion of the feature representations for both model predictor and IoU-Net achieves the best result on VOT-RGBT2019 dataset~\cite{vot2019}. 
We can also see that as pixel-level fusion and response-level fusion both take place in the extra part of the network, they are easier to implement and only fewer variants need to be evaluated, compared with the fusion on the intermediate of the network. 
In the following sections, we select this best performing variant as our final tracker, which we call \emph{mfDiMP}.

\subsection{VOT-RGBT2019 dataset}
In this section, we evaluate our mfDiMP 
on the VOT-RGBT2019 dataset in terms of EAO in Table~\ref{table:VOTRGBT2019}. We compare with several high-quality RGB trackers including ECO~\cite{danelljan2017eco}, ATOM~\cite{danelljan2019atom}, DiMP~\cite{bhat2019learning}, SiameseFC~\cite{bertinetto2016fully}, DaSiamRPN~\cite{zhu2018distractor}. All of these use only the RGB modality as input.
The single modality baseline tracker DiMP~\cite{bhat2019learning}, which shows dominant performances on various RGB datasets~\cite{kristan2018sixth,muller2018trackingnet,fan2018lasot,huang2018got}, also achieves excellent results on VOT-RGBT2019. 
By using our multi-modal fusion with end-to-end training, we improve DiMP by an absolute gain of 6.4\% in terms of EAO.
This significant improvement demonstrates that our selected fusion mechanism is effective for maximally exploiting the multi-modal nature of the given images.

\subsection{RGBT210 dataset}
We evaluate mfDiMP on the recent RGBT210 dataset~\cite{li2017weighted} using their two evaluation metrics (Figure~\ref{fig:RGBT210_plot}). 
We compare against the top-10 trackers on this dataset, including CCOT~\cite{danelljan2016beyond}, ECO~\cite{danelljan2017eco}, CMRT~\cite{li2018cross}, BACF~\cite{kiani2017learning}, SRDCF~\cite{danelljan2015learning}, SGT~\cite{li2017weighted}, Staple~\cite{bertinetto2016staple}, Staple-CA~\cite{mueller2017context}, SiameseFC~\cite{bertinetto2016fully}.
We can see how our tracker significantly outperforms the second best tracker (DiMP) with an absolute gain of 6.7\% and 4.2\%, in terms of precision rate and success rate respectively.
As a result, mfDiMP achieves a new state-of-the-art also on this dataset, bringing further evidence to the advantages of end-to-end training for multi-modal tracking in terms of accurate object localization.
\subsection{Attribute analysis on RGBT210 dataset}
There are a total of 12 different attributes in the RGBT210 dataset~\cite{li2017weighted}.
We analyze the performance of our method on these attributes in terms of precision rate and success rate (PR/SR \%) in Table~\ref{table:RGBT210_attributes}.
We compare with some popular RGB trackers such as ECO~\cite{danelljan2017eco} and CSR~\cite{lukezic2017discriminative}. 
We also compare with the state-of-the-art RGB-T trackers on this dataset, e.g.\ CMRT~\cite{li2018cross} and SGT~\cite{li2017weighted}, and some extended RGB-T trackers~\cite{li2018cross} from RGB modality, e.g. KCF and CFnet. 
This experiment, which compares trackers on specific scenarios, proves the robustness and generality of our mfDiMP on RGB-T tracking.
Our tracker outperforms all other trackers on all attributes but one (thermal crossover). Moreover, in attributes such as partial occlusion, low illumination, deformation, fast motion, camera moving, and background clutter, mfDiMP achieves a significant gain of about 10\% in terms of Success Rate (SR) when compared with the second best.

\section{Conclusions}
\label{conclusion}
Most of the multi-modal trackers are still using hand-crafted features, or simple off-the-shelf deep features. We investigated how to effectively fuse multi-modal data in an end-to-end training manner, which makes optimal use of information from both modalities. We propose three end-to-end multi-modal fusion architectures, consisting of pixel-level fusion, feature-level fusion and response-level fusion. 
To ensure that the proposed fusion tracker can be trained in an end-to-end manner, we also generated a large-scale paired synthetic RGB-T dataset. 
We performed extensive experiments on two recent benchmarks: VOT-RGBT2019~\cite{vot2019} and RGBT210~\cite{li2017weighted}. 
The results showed that the proposed fusion tracker does significantly improve the performance of the baseline tracker with respect to single modality tracking. As a consequence, our end-to-end multi-modal fusion tracker sets new state-of-the-art results on both datasets.
\vspace{5mm}

	\begin{small}
	\noindent\textbf{Acknowledgements.} 
	We acknowledge the financial support by the Spanish project TIN2016-79717-R, and mention the Generalitat de Catalunya CERCA Program.
	\end{small}

{\small
\bibliographystyle{ieee}
\bibliography{shortstrings,egbib}
}

\end{document}